\documentclass[conference]{IEEEtran}
\IEEEoverridecommandlockouts
\usepackage{cite}
\usepackage{amsmath,amssymb,amsfonts}
\usepackage{algorithmic}
\usepackage{graphicx}
\usepackage{textcomp}
\usepackage{xcolor}
\usepackage{cite}
\usepackage{tabularx}
\usepackage{float}
\usepackage{subfig}
\usepackage{url}
\usepackage{tabularx}
	\usepackage{graphicx}
\usepackage{outlines}
\usepackage{enumitem}
\usepackage{csquotes}
\usepackage{amsmath,amssymb,amsfonts}
\usepackage{algorithmic}
\usepackage{graphicx}
\usepackage{textcomp}
\usepackage{xcolor}
\usepackage{pgfplots}
\usepackage{hyperref}
\def\BibTeX{{\rm B\kern-.05em{\sc i\kern-.025em b}\kern-.08em
    T\kern-.1667em\lower.7ex\hbox{E}\kern-.125emX}}
\begin{document}
\title{System Theoretic View on Uncertainties\\
\thanks{The research leading to these results has received funding from the European Union’s Horizon 2020 research and innovation program under the Marie Sklodowska-Curie grant agreement No 812.788 (MSCA-ETN SAS).This publication reflects only the authors view, exempting the European Union from any liability. Project website: \href{url}{http://etn-sas.eu/}.}
}

\author{\IEEEauthorblockN{Roman Gansch}
\IEEEauthorblockA{\textit{Corporate Research,} \\
\textit{Robert Bosch GmbH,}\\
Renningen, Germany \\
first-name.lastname@de.bosch.com}
\and
\IEEEauthorblockN{Ahmad Adee}
\IEEEauthorblockA{\textit{Corporate Research,} \\
	\textit{Robert Bosch GmbH,}\\
	Renningen, Germany \\
	first-name.lastname@de.bosch.com}
}


\maketitle

\begin{abstract}
The complexity of the operating environment and required technologies for highly automated driving is unprecedented. A different type of threat to safe operation besides the fault-error-failure model by Laprie et al. arises in the form of performance limitations. We propose a system theoretic approach to handle these and derive a taxonomy based on uncertainty, i.e. lack of knowledge, as a root cause. Uncertainty is a threat to the dependability of a system, as it limits our ability to assess its dependability properties. We distinguish uncertainties by aleatory (inherent to probabilistic models), epistemic (lack of model parameter knowledge) and ontological (incompleteness of models) in order to determine strategies and methods to cope with them. Analogous to the taxonomy of Laprie et al. we cluster methods into uncertainty prevention (use of elements with well-known behavior, avoiding architectures prone to emergent behavior, restriction of operational design domain, etc.), uncertainty removal (during design time by design of experiment, etc. and after release by field observation, continuous updates, etc.), uncertainty tolerance (use of redundant architectures with diverse uncertainties, uncertainty aware deep learning, etc.) and uncertainty forecasting (estimation of residual uncertainty, etc.).
\end{abstract}

\section{Introduction}
Assuring system safety of highly automated driving vehicles is an unprecedented challenge in engineering. Complex tasks like interpretation of the surrounding world, which previously were handled by a human driver, have to be mastered autonomously by the self-driving car. Novel technologies like machine learning for perception or Lidar sensors are used in these systems. The system environment which is characterized by an open context, i.e. unpredictability of the operational domain \cite{burton2019mind}, and limited knowledge about the achievable performance of the system components results in a high degree of complexity that has to be dealt with.

Established safety engineering practices and standards like ISO2626:2018 \cite{ISO26262:2018} focus on functional safety and well known internal malfunctions of the system. These malfunctions arise due to a deviation of the actual behavior to a well-defined function specification. In complex autonomous systems it is infeasible to provide a complete function specification and fully analyze the implemented behavior for deviations. This leads to multiple gaps in the overall engineering lifecycle that have to be overcome for a safety argumentation \cite{burton2019mind, poddey2019validation}. The root cause for the identified gaps is a lack of knowledge, i.e. uncertainty. This aspect is addressed by the safety of the intended functionality (SOTIF) ISO/PAS21448 standard \cite{ISOPAS21448} in the form of sufficient reduction of unknown unsafe scenarios.

While uncertainty has been considered as a residual risk in the failure of hardware components, it was not incorporated as a general lack of knowledge over the whole engineering lifecycle. Several methods to handle this novel kind of uncertainty have been developed. These include machine learning with uncertainty estimations \cite{gal2016dropout, kendall2017uncertainties} or saliency maps \cite{simonyan2013deep} that can be used for implementation and during runtime. Other aspects of the lifecycle are for example addressed by safety analysis incorporating uncertainty via evidence theory \cite{simon2008bayesian} or verification with probabilistic formal methods \cite{sadigh2016safe, koopman2019autonomous}. For the overall confidence to release a product assurance cases can be enriched with belief modeling \cite{wang2016ds}. These methods provide specialized solutions for problems regarding uncertainty in their respective domain. Some of these methods are complementary, while others are redundant. However, in the overall engineering lifecycle it is not yet clear how these methods should be meshed together to sufficiently address the gaps due to uncertainty.

In this work, we define an initial taxonomy for uncertainties in order to derive this overall strategy. Our work is based on the taxonomy for dependability that has been defined by Laprie et al. \cite{laprie1992dependability, avizienis2004basic}. The taxonomy of Laprie builds on the foundations of system theory and is thus generalized to all application domains. It distinguishes between the properties (availability, reliability, etc.), threats (fault, error, failure) and means to dependability (fault prevention, removal, tolerance and forecasting) and provides a general framework to engineer dependable systems. However, it does not adequately address the impact of uncertainty on dependability, which is a major challenge encountered in autonomous systems and leads to ambiguity when applying the fault-error-failure model. Therefore, we adopt the system theoretic approach to define an appropriate taxonomy. A core element of our taxonomy is a distinction between different types of uncertainty. Analogous to faults, the means to handle these uncertainties are included, which can be used to classify existing methods and derive a holistic strategy.

ncom\section{Systems approach to uncertainty}
In order to formalize the uncertainties encountered in the development of autonomous systems, we adopt the system theoretic viewpoint from cybernetics. We refer to the higly automated driving vehicle being developed as System under Development (SuD). The control loop of interest for us is between the organization developing the SuD (controlling system) and the SuD embedded in its operating environment (controlled system). This viewpoint is analogue to the cognitive system engineering approach by Rasmussen \cite{rasmussen1994cognitive} and the system theoretic accident model and processes (STAMP) by Leveson \cite{leveson2004new}.

\begin{figure}[!htbp]
  \centering
  \includegraphics[width=0.6\columnwidth]{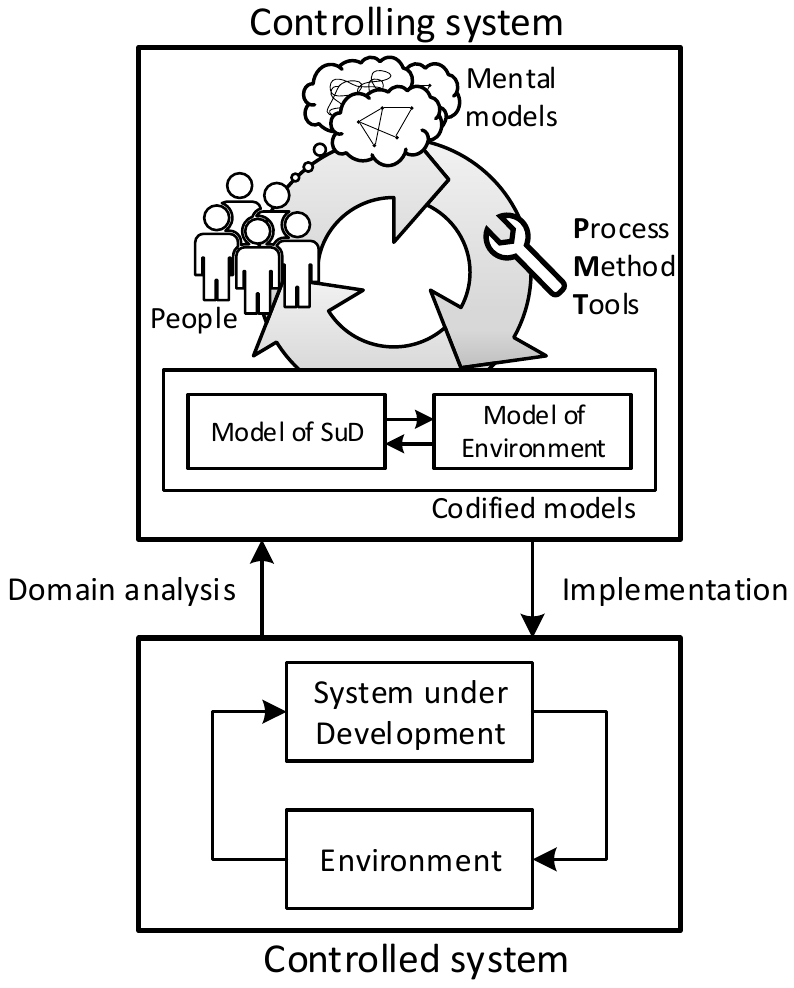}
  \caption{The cybernetic point of view of a development process consists of the controlled system: system under development (SuD) embedded in its environment and the controlling system: development organization.}
  \label{fig:cybernetics}
\end{figure}

The information for developing the SuD is obtained by domain analysis, which acts as an observation channel for the cybernetic control loop (Fig.~\ref{fig:cybernetics}). From the domain analysis developers derive the needs and requirements for the SuD. The developing organization then implements the system and deploys it in the operating environment. The implementation acts as the influencing channel of the control loop. By feedback through prototypes and field observation, the control loop is continuously iterated.

The controlling system exists of several subsystems like people, tools or infrastructure. These are in a complex and symbiotic relationship. For our viewpoint, the detailed interaction are not of interest. We are mostly interested in the mental and codified models within the organization. The mental models are formed on an individual basis and represent the state of knowledge of each person, which also contains tacit knowledge, i.e. knowledge that cannot be codified into information \cite{reber1989implicit}. The individual knowledge is made explicit by transformation into information within the codified models by using processes, methods and tools. The codified models are used for several purposes like communication of mental models between developers or implementation of the system.

A fundamental principle of the cybernetic control loop is the good regulator theorem formulated by Ashby \cite{ashby1957introduction, conant1970every}. It states that every good regulator of a system must be a model of that system in order to achieve a good outcome (i.e. a SuD fulfilling its purpose). The controlling system needs to be able to predict the influence of its controlling actions on the controlled system with a model. This shows the importance of the models, as these have to be accurate to perform the correct actions. Uncertainty can be attributed to inaccurate or wrong model.

\subsection{System models}
\begin{figure}[!htbp]
  \centering
  \includegraphics[width=1\columnwidth]{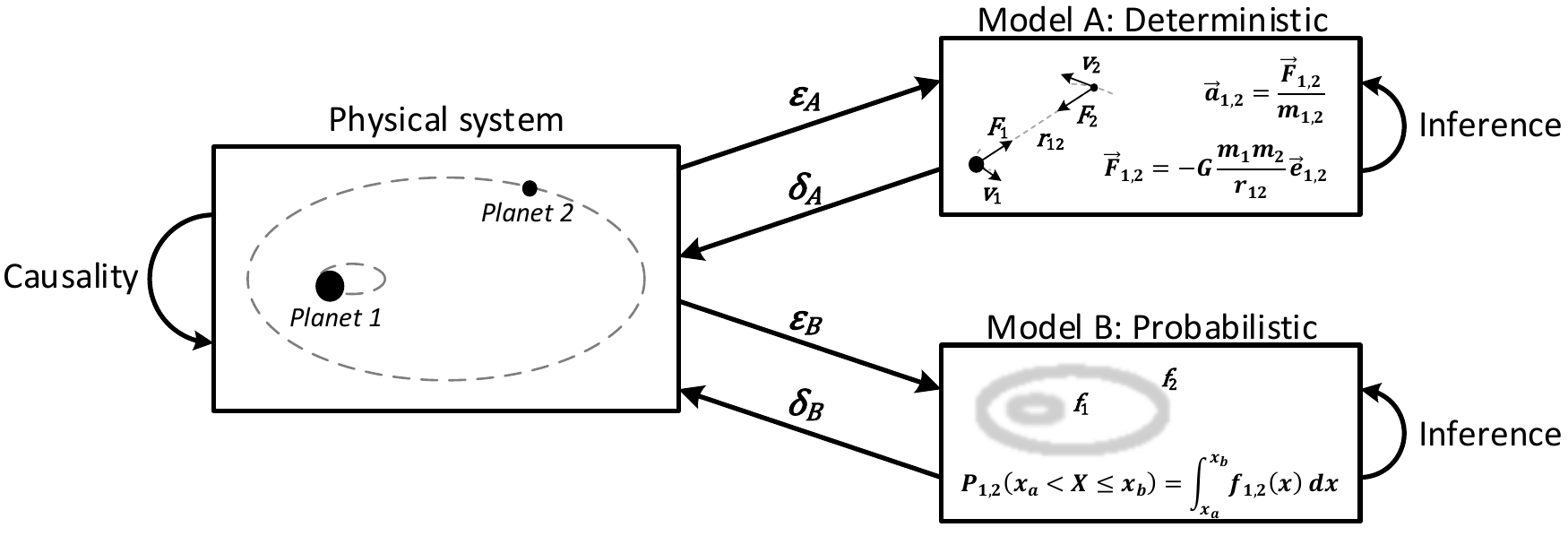}
  \caption{Modeling relation between a physical system consisting of two ideal point masses (planet 1 and 2) and two formal systems as models.}
  \label{fig:modelling_relation}
\end{figure}

It is crucial to understand what a model is composed of and which properties it has to fulfill in order to be useful. A formal basis for models has been formulated by Rosen \cite{rosen1991life}. He stresses the importance of the modeling relation (Fig.~\ref{fig:modelling_relation}): an isomorphic encoding $\epsilon_{A,B}$ and decoding $\delta_{A,B}$ of relevant properties of the physical system into formal systems. The causality in the physical system is thereby mapped to logic inferences in the model. Such models can be mathematical equations, SysML diagrams or, in the context of safety engineering, failure analysis models (e.g. fault tree analysis).

Multiple modeling relations between physical and formal systems can be established, i.e. there are multiple models available for a system. These are different representations tailored to the specific needs of the modeler. Since our focus is on the uncertainty aspect of modeling, it is important to distinguish between deterministic and probabilistic models. From the former a singular outcome can be inferred for a given input, while for the latter only statements about probabilistic outcomes can be inferred. On a very fine granular level the physical world itself becomes inherently probabilistic due to quantum mechanics. For these physical processes only probabilistic models are applicable. However, on a macroscopic level deterministic relations dominate causality and it is the choice of the modeler to choose either a deterministic or probabilistic representation of a physical system as a model.

To give an example, consider a reality where only two planets exist, with a homogeneous mass distribution (i.e. they can be treated as point masses). These exert a gravitational force on each other and influence each other's movement through space and time resulting in elliptic orbits. We will show how a deterministic as well as probabilistic model can be built for the same physical system.

The behavior of this system can be described by Newton's laws (Fig.~\ref{fig:modelling_relation}: model A). The mathematical formulas for gravitational forces and motion result in a set of differential equations. These represent a deterministic model that allows inferring the causality of motion of the bodies. Given the starting conditions of initial position and velocity, every future state of the system can be inferred. In principle, it would be possible to also apply this to arbitrarily complex many-body problems. Regardless of the complexity of the problem, calculating the future state of the entire universe is always possible with a hypothetical Laplace demon \cite{laplace1998pierre}. An entity that knows the entire present state of the universe with unlimited calculation capacity to exactly predict the future state. In reality, however there are practical (availability of information, computation capacity, etc.) as well as theoretical (G\"odel's incompleteness theorem, quantum mechanics, free will, etc.) limits to determinism \cite{earman1986primer}.

Another way to describe the system of two planets is to adopt the frequentist point of view (Fig. \ref{fig:modelling_relation}: model B). This means, to build a probabilistic model by repeated observation of the positions. With an infinite amount of observations, the exact probabilities to find either of the two bodies within a spatial frame can be inferred from the obtained probability density functions. 

Both models, the deterministic and probabilistic, fulfill the modeling relation and allow to draw conclusions about the behavior of the system. Each model has its own purpose and is valid for a given set of behavior that the modeler wants to describe. While the deterministic one is useful to e.g. calculate the trajectory for satellites, the probabilistic one is useful e.g. for predicting the chance of meteor impacts.

This example illustrates that the property of determinism and probability also have to be attributed to the model and not only to the system itself. Only when dealing with systems that are strongly influenced by the theoretical limits of determinism the probabilistic nature becomes inherent. Models can also be a mixture of deterministic and probabilistic elements and it is the choice of the modeler, which representation of the system best fits his needs.


\section{Types of Uncertainties}
Uncertainties refer to the notion of unknown or insufficient knowledge \cite{hastings2004framework}. As a generic concept of the unknown, incomplete or imperfect knowledge, it has been classified in various contexts in the literature \cite{walley1991statistical,clarkson2010design,de2007classification,der2009aleatory}. 
In our system theoretic approach to developing highly automated driving vehicles, we focus on the importance of models to represent our knowledge. Depending upon the origins of uncertainties in our models, we distinguish between aleatory, epistemic and ontological uncertainties. 

\subsection{Aleatory Uncertainty}
\emph{Aleatory uncertainty can be regarded as randomness of a process represented by a system model.}

Aleatory uncertainty is considered to be irreducible for a given choice of a probabilistic model and is quantified by probabilistic distributions \cite{der2009aleatory}. 

As an example, we continue with the physical system consisting of two planets of the previous section. We are interested in the probabilistic spatial distribution of the point masses. We therefore use model B of Fig.~\ref{fig:modelling_relation}, which provides a probability that a planet is found in a given spatial frame. With this probabilistic model of spatial distribution, the probabilities to find a point mass in a certain frame depict a mere belief and the crisp location at a given instant is subject to aleatory uncertainty. 
   
\subsection{Epistemic Uncertainty}
\emph{Epistemic uncertainty is associated to lack of knowledge about the system model and the inexact encoding of physical system to models.}

An important aspect of epistemic uncertainty is that we lack in knowledge but we are aware of it. This can also be referred to as the \emph{known-unknown} of the model \cite{taleb2007black}. Another characterization of epistemic uncertainty can be attributed to conditional entropy \cite{shannon1948mathematical, heylighen2001cybernetics}, i.e. the difference of information contained in the model with respect to the physical system. A model is an abstracted encoding of a system, hence a mere projection of reality. Due to the abstraction, not every detail is encoded in the model and thereby conditional entropy is induced.

We extend the concept of epistemic uncertainty to our universe of point masses and start with the deterministic model first. This model allows exact calculation of the motion of the planets. For the idealized point masses the model is completely accurate and there is no uncertainty in this model. However, real world scenarios are more complex than point masses. Now consider the planets with a homogeneous mass distribution are replaced by a heterogeneous body with an uneven surface. Newton's laws still hold, but a model with idealized point masses does not accurately describe the physical system anymore. The causality in the physical system are not completely encoded into the formal system model. The lack of knowledge due to the inaccurate model leads to an epistemic uncertainty. A more accurate model based on a spatial integral over the attracting forces would allow a more accurate description and reduce the epistemic uncertainty. However, this would require complete information of the actual mass distribution, which might not be available.

Now consider the probabilistic model representation. The probabilistic distribution of the point masses is an aleatory uncertainty for the given model. However, the probabilities can usually not be determined exactly. With a frequentist approach, we calculate the probabilities based on our past observations. The gap between the actual and observed probabilities represents the epistemic uncertainty of the probabilistic model. With each new observation, our distribution parameter (e.g. means and variance for a Gaussian distribution) become more credible. Hence, our knowledge increases and the epistemic uncertainty decreases with every observation.

\subsection{Ontological Uncertainty}
\label{onto}
\emph{Ontological uncertainty can be defined as a condition of complete ignorance in the model of a relevant aspect of the system.}

This can also be referred to as the \emph{unknown-unknown} \cite{taleb2007black}, the state of we do not know that we do not know. The term ontological uncertainty is based on ontology, i.e. is the study of existence. We thereby refer to a lack of knowledge about the actual existence of relevant aspects in our model representation.

Again, we use the point mass universe to explain the concept of uncertainty. We assumed that there are only two planets and our past experience with the universe also supports this hypothesis. However, at some point we observe a behavior of the planets that contradicts the prediction by the models due to the influence of a third planet. This challenges our prior beliefs about the universe. The probabilistic and deterministic model are completely inaccurate. The possibility that there are phenomena in the universe which are neglected by our models is represented by the ontological uncertainty. In order to formulate a new model that provide the necessary representation of the universe, we need to reformulate our models to include the influence of the third point mass.

Previous work only distinguishes between aleatory and epistemic uncertainty \cite{der2009aleatory}. However, we consider that in the field of highly automated driving vehicles, a further distinction into ontological uncertainty is valuable, as they require different means for reduction. The completeness and correctness of models are generally challenged in the early stages of research and development of novel systems. For autonomous systems operating in an open context the ontological uncertainty can never be completely disregarded. Therefore, it should be included in the safety case that it has been properly addressed.

Segregating the epistemic and ontological uncertainty in a model can be challenging, as they constitute of blurred boundaries. A general rule of thumbs is a distinction between model accuracy (epistemic) and model correctness (ontological). Subjectively these uncertainties can be distinguished by the surprise factor when we observe new behavior. Mathematically the conditional entropy between the system and its model can be used as a formal expression for the surprise factor \cite{shannon1948mathematical, heylighen2001cybernetics}. 
\section{Means of Uncertainties}
\label{means}
\begin{figure}[!htbp]
	\centering
	\includegraphics[width=0.8\columnwidth]{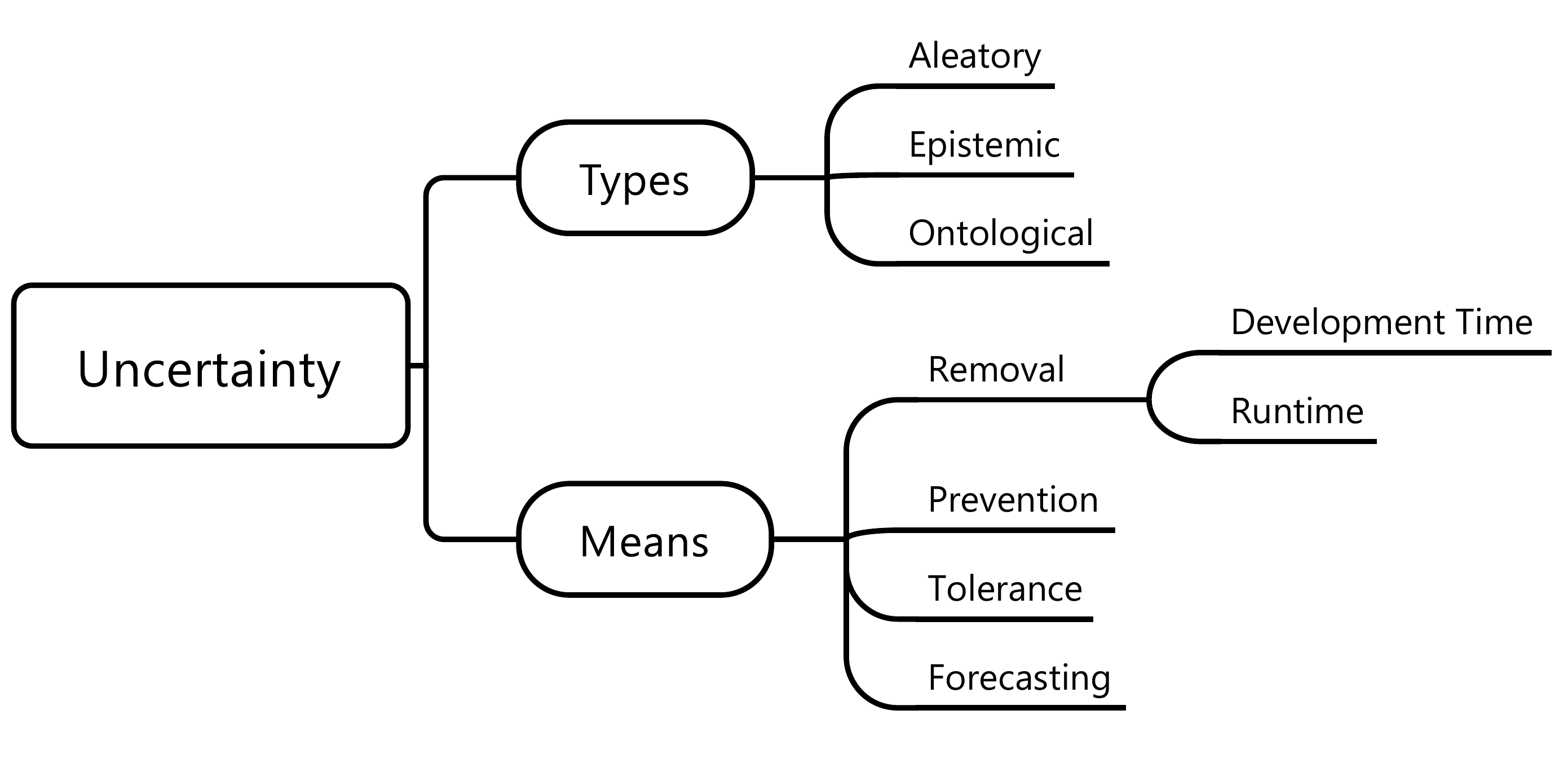}
	\caption{Uncertainty occurring in system models are classified into aleatory, epistemic and ontological. To build dependable system in the presence of these uncertainties the means prevention, tolerance, removal and/or forecasting are necessary.}
	\label{fig:Means}
\end{figure}

In order to develop dependable systems, the impact of aleatory, epistemic and ontological uncertainties on the overall dependability has to be minimized. Analogous to the taxonomy for dependability given by Laprie et al. \cite{laprie1992dependability, avizienis2004basic}, we use a similar taxonomy for means to cope with uncertainties (Fig.~\ref{fig:Means}).

In this uncertainty taxonomy, we classify the means into uncertainty prevention, tolerance, removal and forecasting:
\begin{itemize}
	\item Uncertainty prevention can e.g. be achieved by avoiding complexity in the system. This can be done by using simple architectures not prone to emergent behavior or restriction of the operational design domain (ODD).
	\item Uncertainty removal can be done during development by e.g. a safety analysis including epistemic/ontological uncertainty or during use by e.g. field observation to monitor ontological events.
	\item Uncertainty tolerance can typically be obtained by using redundant architectures (e.g. overlapping field of views of sensors) or using components that can detect uncertainty (e.g. machine learning with epistemic uncertainty outputs).
	\item Uncertainty forecasting is based on estimating the present level and future occurrence of uncertainties. These are relevant to make a decision about the release of a product by e.g. arguing about a sufficiently low ontological uncertainty. 
\end{itemize}

This taxonomy can be used to define a strategy to deal with uncertainty on system and component level. As a general rule of thumb, uncertainty prevention should be prioritized as this eliminates the need for further measures. Uncertainty removal should be especially considered in design processes to identify weaknesses in the architecture. However, due to the open context it will not be possible to sufficiently reduce uncertainty by only focusing on prevention and removal. Uncertainty tolerance within the system is required to cope with unforeseen scenarios.

In order to systematically address the gaps due to uncertainty, we aim to use the different types to evaluate the capability of various means. Especially ontological uncertainty is difficult handle and is the main contributor to the long tail validation challenge of highly automated driving \cite{longtailA, koopman2018heavy}. Methods like uncertainty tolerance are hardly able to cope with this type, due to the limit of knowledge that can be incorporated into the implemented system. Instead, methods like uncertainty removal during use are better suited. To demonstrate how the different types of uncertainty can be addressed by the means, we explore uncertainty removal as an example by extending a safety analysis with uncertainty in the next section.
\section{Uncertainty removal}
\label{Safety Assurance}
Safety analysis are an integral part of safety engineering to verify that the design fulfills desired properties. An established method is the fault tree analysis (FTA) \cite{osti_382787}. In this section, we first give a brief introduction to FTA. It can be challenging to include epistemic and ontological uncertainties explicitly within the FTA's failure model. By using more sophisticated analysis models, we show that we can include these types of uncertainty. 

\subsection{Fault tree analysis}
\label{Fault Tree Analysis}
FTA is a graphical model based on a Boolean fault propagation and is used to identify shortcomings like single point faults in the system. Different extensions to the basic method have been introduced to deal with more complex aspects of analysis \cite{dugan1992dynamic,tanaka1983}. As a deductive analysis tool, FTA starts with identifying undesirable top events. This top down process is continued until root causes (basic events) are identified which cannot be further decomposed. The propagation of basic events to the undesired top event is modeled by Boolean gates.

While FTA is quite popular and in widespread uses, it also has some shortcomings. For autonomous systems, the failure oriented nature of FTA limits the ability to include human factors or nominal performance of the system. Further, the cause and effect relationship between events is deterministic, which cannot model more diverse and uncertain relations in the systems. Advances to model this probabilistically in FTA models have been made \cite{ferdous2011fault}. However, these still contain several limitations to be able to model more generalized dependencies and uncertain relationships which are encountered in complex systems. 

\subsection{Safety analysis with uncertainty}
\label{PRA}
In order to deal with the shortcomings of FTA regarding the different types of uncertainty we propose an analysis method based on evidence theory \cite{shafer1976mathematical} in combination with Bayesian networks (BN) \cite{simon2008bayesian}. Evidence theory incorporates the different types of uncertainty and the Bayesian network representation allows to use a graphical model that supports in construction of the mental model by the developers.

The BN is a Directed Acyclic Graph (DAG) that consists of nodes and edges. Every node is a random variable, which represents an element of the system or its context. The edges represent a directed causal relationship between two nodes. The edge direction runs from the parent node (pa) towards the child node (ch). Together, node and edge represent the structure of the probabilistic network. The effect of parent node on child node is determined by conditional probabilities $P(pa|ch)$. In the following, we demonstrate how all three types of uncertainties can be modeled in a BN with a simple example.

\begin{figure}
	\centering
	\includegraphics[width=0.85\columnwidth]{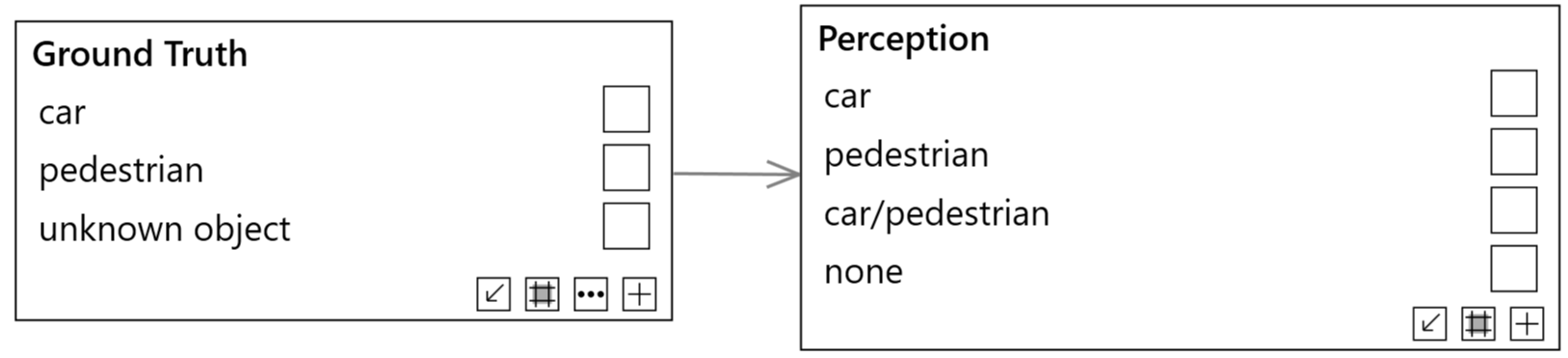}
	\caption{Bayesian Network of an object perception chain. The ground truth node represents the actual object and the perception node the classified object as output.}
	\label{fig:BNN}
\end{figure}

Consider we want to develop a perception chain consisting of a camera with a machine learning algorithm that classifies objects. When developing this system we assume that only cars or pedestrians will be encountered, and hence only these will be included in the model used by the perception chain. However, we also believe in the possibility of existence of other objects that will be encountered in the real world and include an unknown object state in our root node (Fig.~\ref{fig:BNN}). This unknown object state of the ground truth represents the ontological uncertainty in our analysis. The probabilities $P_{car} = 0,6$, $P_{ped} = 0,3$, $P_{unknown} = 0,1$ represent how likely we will encounter these objects in the real world. This probabilistic distribution represents the aleatory uncertainty of our world model.

At the end of the perception chain, we classify the encountered objects to car, pedestrian or none. We also include a state about car/pedestrian. This state is not an actual output of the classifier. It represents our epistemic uncertainty in assessing the performance of the perception chain to correctly classify the objects. The conditional probability table (CPT, Tab.~\ref{tab:14}) of the perception node in the BN encodes this epistemic uncertainty by the corresponding probabilities in the car/pedestrian column.

\begin{table}
\caption{Conditional probability table (CPT) of $P(perception|groundtruth)$. Epistemic uncertainty about the performance of the classification is encoded in the probabilities of the state car/pedestrian. Ontological uncertainty about the possibility of further objects is included in the unknown state of the ground truth.}
\label{tab:14}
\begin{tabular}{|c|cccc|}
\hline
& \multicolumn{4}{c|}{Perception}\\
Ground Truth    & car	& pedestrian & car/pedestrian & none\\
\hline
car				& 0.9 	& 0.005 & 0.05 & 0.045\\
pedestrian  	& 0.005 & 0.9	& 0.05 & 0.045\\
unknown		    & 0 	& 0  	& 0.2  & 0.7\\
\hline
\end{tabular}
\end{table}

The advantage of the BN approach is that it can be scaled up to model the complete system and allows hierarchical refinement analogous to FTA. For each node and CPT the corresponding aleatory, epistemic and ontological uncertainty can be included as required. However, the number of parameters that need to be elicited in the CPT grows exponentially with the number of parent nodes and their states. Using BN for larger systems, hence, can become cumbersome. However, several techniques to deal with this problem are available \cite{fenton2007using,chin2009assessing,mendes2018towards}.

In this example, we have shown a clear distinction between epistemic uncertainty (decision between existing states) and ontological uncertainty (objects not included in the perception model). This distinction is essential for uncertainty removal, as it allows identification of fitting measures. The epistemic uncertainty can be reduced by further observation and refinement of the existing perception models (i.e. increase of our knowledge about the perception chain performance). In contrast, the ontological uncertainty requires a more thorough domain analysis and extension of the perception model.

With this kind of analysis, it can also be demonstrated that redundant architectures with diverse uncertainties can be used to build uncertainty tolerant systems. The BN approach also allows including dependencies by common parent nodes to identify common causes for uncertainties.

\section {Conclusion and outlook}
\label{futurework}
We presented a system theoretic viewpoint of uncertainty and derived a distinction between aleatory, epistemic and ontological types. Due to the different nature of these types, they have to be addressed by fitting methods. In order to provide a framework to define an overall strategy for methods suited to deal with these uncertainties, we classify the means into prevention, removal, tolerance and forecasting. As an example for applying the types of uncertainty, we have shown a safety analysis method by using a Bayesian network approach with evidence theory which can be used for uncertainty removal.

We intend to apply this system-theoretic viewpoint to all steps of development process and use it as guidance to decide which means are best suited to handle the three types of uncertainty. Thereby, we want to build a safety argument that uncertainties are properly managed and do not pose an unacceptable level of risk.

\bibliographystyle{IEEEtran}
\bibliography{mybibfile}

\begin{thebibliography}{10}
\providecommand{\url}[1]{#1}
\csname url@samestyle\endcsname
\providecommand{\newblock}{\relax}
\providecommand{\bibinfo}[2]{#2}
\providecommand{\BIBentrySTDinterwordspacing}{\spaceskip=0pt\relax}
\providecommand{\BIBentryALTinterwordstretchfactor}{4}
\providecommand{\BIBentryALTinterwordspacing}{\spaceskip=\fontdimen2\font plus
\BIBentryALTinterwordstretchfactor\fontdimen3\font minus
  \fontdimen4\font\relax}
\providecommand{\BIBforeignlanguage}[2]{{%
\expandafter\ifx\csname l@#1\endcsname\relax
\typeout{** WARNING: IEEEtran.bst: No hyphenation pattern has been}%
\typeout{** loaded for the language `#1'. Using the pattern for}%
\typeout{** the default language instead.}%
\else
\language=\csname l@#1\endcsname
\fi
#2}}
\providecommand{\BIBdecl}{\relax}
\BIBdecl

\bibitem{burton2019mind}
S.~Burton, I.~Habli, T.~Lawton, J.~McDermid, P.~Morgan, and Z.~Porter, ``Mind
  the gaps: Assuring the safety of autonomous systems from an engineering,
  ethical, and legal perspective,'' \emph{Artificial Intelligence}, p. 103201,
  2019.

\bibitem{ISO26262:2018}
{International Organization for Standardization}, ``{ISO26262: Road vehicles -
  Functional safety},'' {2018}.

\bibitem{poddey2019validation}
A.~Poddey, T.~Brade, J.~E. Stellet, and W.~Branz, ``On the validation of
  complex systems operating in open contexts,'' \emph{arXiv preprint
  arXiv:1902.10517}, 2019.

\bibitem{ISOPAS21448}
{International Organization for Standardization}, ``{ISO/PAS21448: Road
  vehicles - Safety of the intended functionality},'' {2019}.

\bibitem{gal2016dropout}
Y.~Gal and Z.~Ghahramani, ``Dropout as a bayesian approximation: Representing
  model uncertainty in deep learning,'' in \emph{International conference on
  machine learning}, 2016, pp. 1050--1059.

\bibitem{kendall2017uncertainties}
A.~Kendall and Y.~Gal, ``What uncertainties do we need in bayesian deep
  learning for computer vision?'' in \emph{Advances in neural information
  processing systems}, 2017, pp. 5574--5584.

\bibitem{simonyan2013deep}
K.~Simonyan, A.~Vedaldi, and A.~Zisserman, ``{Deep inside convolutional
  networks: Visualising image classification models and saliency maps},''
  \emph{arXiv preprint arXiv:1312.6034}, 2013.

\bibitem{simon2008bayesian}
C.~Simon, P.~Weber, and A.~Evsukoff, ``{Bayesian networks inference algorithm
  to implement Dempster Shafer theory in reliability analysis},''
  \emph{Reliability Engineering \& System Safety}, vol.~93, no.~7, pp.
  950--963, 2008.

\bibitem{sadigh2016safe}
D.~Sadigh and A.~Kapoor, ``Safe control under uncertainty with probabilistic
  signal temporal logic,'' 2016.

\bibitem{koopman2019autonomous}
P.~Koopman, B.~Osyk, and J.~Weast, ``Autonomous vehicles meet the physical
  world: Rss, variability, uncertainty, and proving safety,'' in
  \emph{International Conference on Computer Safety, Reliability, and
  Security}.\hskip 1em plus 0.5em minus 0.4em\relax Springer, 2019, pp.
  245--253.

\bibitem{wang2016ds}
R.~Wang, J.~Guiochet, G.~Motet, and W.~Sch{\"o}n, ``{DS} theory for argument
  confidence assessment,'' in \emph{International Conference on Belief
  Functions}.\hskip 1em plus 0.5em minus 0.4em\relax Springer, 2016, pp.
  190--200.

\bibitem{laprie1992dependability}
J.-C. Laprie, ``{Dependability: Basic concepts and terminology},'' in
  \emph{Dependability: Basic Concepts and Terminology}.\hskip 1em plus 0.5em
  minus 0.4em\relax Springer, 1992, pp. 3--245.

\bibitem{avizienis2004basic}
A.~Avizienis, J.-C. Laprie, B.~Randell, and C.~Landwehr, ``Basic concepts and
  taxonomy of dependable and secure computing,'' \emph{IEEE transactions on
  dependable and secure computing}, vol.~1, no.~1, pp. 11--33, 2004.

\bibitem{rasmussen1994cognitive}
J.~Rasmussen, A.~M. Pejtersen, and L.~P. Goodstein, ``Cognitive systems
  engineering,'' 1994.

\bibitem{leveson2004new}
N.~Leveson, ``A new accident model for engineering safer systems,''
  \emph{Safety science}, vol.~42, no.~4, pp. 237--270, 2004.

\bibitem{reber1989implicit}
A.~S. Reber, ``Implicit learning and tacit knowledge.'' \emph{Journal of
  experimental psychology: General}, vol. 118, no.~3, p. 219, 1989.

\bibitem{ashby1957introduction}
W.~R. Ashby, ``An introduction to cybernetics,'' 1957.

\bibitem{conant1970every}
R.~C. Conant and W.~Ross~Ashby, ``Every good regulator of a system must be a
  model of that system,'' \emph{International journal of systems science},
  vol.~1, no.~2, pp. 89--97, 1970.

\bibitem{rosen1991life}
R.~Rosen, \emph{Life itself: a comprehensive inquiry into the nature, origin,
  and fabrication of life}.\hskip 1em plus 0.5em minus 0.4em\relax Columbia
  University Press, 1991.

\bibitem{laplace1998pierre}
P.-S. Laplace, \emph{Pierre-Simon Laplace Philosophical Essay on Probabilities:
  Translated from the fifth French edition of 1825 With Notes by the
  Translator}.\hskip 1em plus 0.5em minus 0.4em\relax Springer Science \&
  Business Media, 1998, vol.~13.

\bibitem{earman1986primer}
J.~Earman \emph{et~al.}, \emph{A primer on determinism}.\hskip 1em plus 0.5em
  minus 0.4em\relax Springer Science \& Business Media, 1986, vol.~37.

\bibitem{hastings2004framework}
D.~Hastings and H.~McManus, ``A framework for understanding uncertainty and its
  mitigation and exploitation in complex systems,'' in \emph{2004 Engineering
  Systems Symposium}, 2004, pp. 29--31.

\bibitem{walley1991statistical}
P.~Walley, \emph{Statistical reasoning with imprecise probabilities}.\hskip 1em
  plus 0.5em minus 0.4em\relax London New York: Chapman and Hall, 1991.

\bibitem{clarkson2010design}
J.~Clarkson and C.~Eckert, \emph{Design process improvement: a review of
  current practice}.\hskip 1em plus 0.5em minus 0.4em\relax Springer Science \&
  Business Media, 2010.

\bibitem{de2007classification}
O.~De~Weck, C.~M. Eckert, P.~J. Clarkson \emph{et~al.}, ``A classification of
  uncertainty for early product and system design,'' in \emph{DS 42:
  Proceedings of ICED 2007, the 16th International Conference on Engineering
  Design, Paris, France, 28.-31.07. 2007}, 2007, pp. 159--160.

\bibitem{der2009aleatory}
A.~Der~Kiureghian and O.~Ditlevsen, ``Aleatory or epistemic? does it matter?''
  \emph{Structural Safety}, vol.~31, no.~2, pp. 105--112, 2009.

\bibitem{taleb2007black}
N.~N. Taleb, \emph{The black swan: The impact of the highly improbable}.\hskip
  1em plus 0.5em minus 0.4em\relax Random house, 2007, vol.~2.

\bibitem{shannon1948mathematical}
C.~E. Shannon, ``A mathematical theory of communication,'' \emph{Bell system
  technical journal}, vol.~27, no.~3, pp. 379--423, 1948.

\bibitem{heylighen2001cybernetics}
F.~Heylighen and C.~Joslyn, ``Cybernetics and second-order cybernetics,''
  \emph{Encyclopedia of physical science \& technology}, vol.~4, pp. 155--170,
  2001.

\bibitem{longtailA}
``{Autonomous Vehicles vs. Kangaroos: the Long Furry Tail of Unlikely
  Events},''
  {https://spectrum.ieee.org/cars-that-think/transportation/self-driving/autonomous-cars-vs-kangaroos-the-long-furry-tail-of-unlikely-events},
  accessed: 2019-11-28.

\bibitem{koopman2018heavy}
P.~Koopman, ``The heavy tail safety ceiling,'' in \emph{Automated and Connected
  Vehicle Systems Testing Symposium}, 2018.

\bibitem{osti_382787}
H.~Barringer, ``An overview of reliability engineering principles,'' 11 1996.

\bibitem{dugan1992dynamic}
J.~B. Dugan, S.~J. Bavuso, and M.~A. Boyd, ``Dynamic fault-tree models for
  fault-tolerant computer systems,'' \emph{IEEE Transactions on reliability},
  vol.~41, no.~3, pp. 363--377, 1992.

\bibitem{tanaka1983}
H.~{Tanaka}, L.~T. {Fan}, F.~S. {Lai}, and K.~{Toguchi}, ``Fault-tree analysis
  by fuzzy probability,'' \emph{IEEE Transactions on Reliability}, vol. R-32,
  no.~5, pp. 453--457, Dec 1983.

\bibitem{ferdous2011fault}
R.~Ferdous, F.~Khan, R.~Sadiq, P.~Amyotte, and B.~Veitch, ``Fault and event
  tree analyses for process systems risk analysis: uncertainty handling
  formulations,'' \emph{Risk Analysis: An International Journal}, vol.~31,
  no.~1, pp. 86--107, 2011.

\bibitem{shafer1976mathematical}
G.~Shafer, \emph{A mathematical theory of evidence}.\hskip 1em plus 0.5em minus
  0.4em\relax Princeton university press, 1976, vol.~42.

\bibitem{fenton2007using}
N.~E. Fenton, M.~Neil, and J.~G. Caballero, ``Using ranked nodes to model
  qualitative judgments in bayesian networks,'' \emph{IEEE Transactions on
  Knowledge and Data Engineering}, vol.~19, no.~10, pp. 1420--1432, 2007.

\bibitem{chin2009assessing}
K.-S. Chin, D.-W. Tang, J.-B. Yang, S.~Y. Wong, and H.~Wang, ``Assessing new
  product development project risk by bayesian network with a systematic
  probability generation methodology,'' \emph{Expert Systems with
  Applications}, vol.~36, no.~6, pp. 9879--9890, 2009.

\bibitem{mendes2018towards}
E.~Mendes, P.~Rodriguez, V.~Freitas, S.~Baker, and M.~A. Atoui, ``Towards
  improving decision making and estimating the value of decisions in
  value-based software engineering: the value framework,'' \emph{Software
  Quality Journal}, vol.~26, no.~2, pp. 607--656, 2018.

\end{thebibliography}

\end{document}